\documentclass[a4paper,conference]{IEEEtran}
\usepackage{times}
\usepackage{graphicx}
\usepackage{amsmath}
\usepackage{amssymb}
\usepackage{cite}
\usepackage{multirow}
\usepackage{color}
\usepackage{bm}
\usepackage{enumitem}
\usepackage{hyperref} 
\usepackage{wrapfig}

\ifCLASSINFOpdf
\else
\fi
\begin{document}
%
\title{Learn From the Past: Experience Ensemble Knowledge Distillation}

\author{\IEEEauthorblockN{Chaofei Wang\textsuperscript{1},
		Shaowei Zhang\textsuperscript{2},
		Shiji Song\textsuperscript{1},
		Gao Huang\textsuperscript{1}}\\
	\IEEEauthorblockA{$^{1}$Department of Automation, Tsinghua University, Beijing, China wangcf18@mails.tsinghua.edu.cn}\\$^{2}$Tianjin University, Tianjin, China sw\_zhang@tju.edu.cn}


%


\maketitle

\begin{abstract}
Traditional knowledge distillation transfers ``dark knowledge'' of a pre-trained teacher network to a student network, and ignores the knowledge in the training process of the teacher, which we call teacher's experience. However, in realistic educational scenarios, learning experience is often more important than learning results. In this work, we propose a novel knowledge distillation method by integrating the teacher's experience for knowledge transfer, named experience ensemble knowledge distillation (EEKD). We save a moderate number of intermediate models from the training process of the teacher model uniformly, and then integrate the knowledge of these intermediate models by ensemble technique. A self-attention module is used to adaptively assign weights to different intermediate models in the process of knowledge transfer. Three principles of constructing EEKD on the quality, weights and number of intermediate models are explored. A surprising conclusion is found that strong ensemble teachers do not necessarily produce strong students. The experimental results on CIFAR-100 and ImageNet show that EEKD outperforms the mainstream knowledge distillation methods and achieves the state-of-the-art. In particular, EEKD even surpasses the standard ensemble distillation on the premise of saving training cost.



\end{abstract}


%
\IEEEpeerreviewmaketitle

\section{Introduction}

In the past few years, deep neural networks (DNNs) have brought excellent performance in many visual classification tasks. Facing the increasingly complex data, DNNs have continuously been improved with more complicated structures (AlexNet\cite{krizhevsky2012imagenet}, VGGNet\cite{simonyan2014very}, ResNet\cite{he2016deep}, ResNext\cite{xie2017aggregated}, DenseNet\cite{huang2017densely}, ViT\cite{dosovitskiy2020image}), which means massive parameters and expensive computation and storage cost. However, it is difficult to deploy such large networks on resource-limited embedded systems. Along with the increasing demands for low-cost networks running on embedded systems, there is an urgency for getting smaller networks with less computation and memory cost, while narrowing the gap of performance between small and large networks.

\begin{figure}[t]
	\setlength{\belowcaptionskip}{-1.cm}
	\begin{center}
		\includegraphics[width=0.99\linewidth]{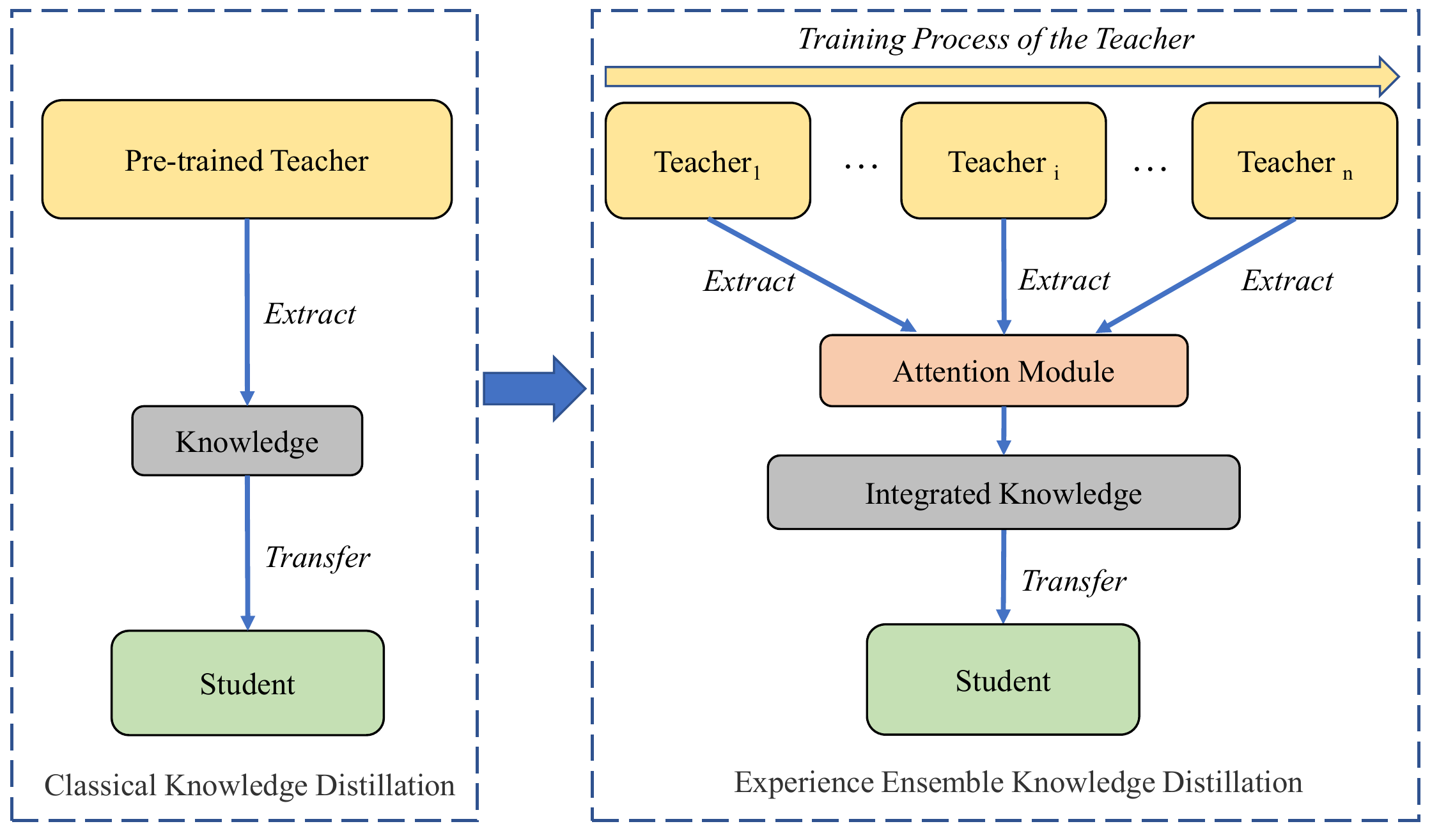}
	\end{center}
	\caption{The sketch map of experience ensemble knowledge distillation (EEKD) in the teacher-student framework. It extracts the knowledge of the intermediate models from the teacher's training process, integrate the knowledge through a self-attention module, and then transfers the knowledge to the student network.}
	\label{fig:intuition}
\end{figure}
Several techniques have been proposed to address this issue, e.g., parameter pruning and sharing \cite{han2015learning,molchanov2016pruning}, compact convolutional filters \cite{zhang2018shufflenet,howard2017mobilenets}, low-rank factorization \cite{denton2014exploiting,jaderberg2014speeding} and knowledge distillation \cite{hinton2015distilling}. Among these routes, knowledge distillation has been proved as an effective way to promote the performance of small network by transferring some ``dark knowledge" from a large teacher model. Once trained, this small network can be directly deployed on resource-limited devices. The key problem of knowledge distillation is how to represent the knowledge of the teacher. A vanilla knowledge distillation \cite{hinton2015distilling} uses the soft targets (the outputs of the final softmax function with a temperature factor) as the teacher knowledge. Subsequently, other kinds of knowledge such as hints\cite{romero2014fitnets}, attention maps\cite{zagoruyko2016paying,wang2021towards}, probability distribution of features\cite{passalis2018learning}, relationships between samples\cite{peng2019correlation} or layers\cite{yim2017gift} are proposed. However, they all focus on the knowledge formed after the training of the teacher, and ignore the valuable knowledge during the training process of the teacher, which is named \emph{teacher's experience}. In the case of people, an excellent teacher should not only tell the students what is the truth, but also tell them his experience in the pursuit of the truth. Although his experience may include failures and detours, but it is even more important than the truth itself. 

Based on this motivation, we propose a novel knowledge distillation method, named experience ensemble knowledge distillation (EEKD), which aims to integrate and transfer the teacher's experience knowledge to the student network. Fig.\ref{fig:intuition} shows the sketch map of EEKD. Specifically, we first train the teacher network, uniformly preserving an appropriate number of intermediate models during the training, and then integrate the knowledge of these intermediate models by ensemble technique, finally transfer the knowledge to the student network. In particular, a self-attention module is used to adaptively assign weights to different intermediate models.  

Furthermore, we explore the principles of constructing EEKD from three aspects, including the quality, weights and number of intermediate models. We find these conclusions: 1) High-performance EEKD does not depend on high-quality intermediate models. 2) Attention-based weight strategy is better than mean, linear increase and linear decrease. 3) Trade-off between performance and cost should be considered when setting the number of intermediate models. Further analysis, a deeper conclusion is that \emph{a high-performance ensemble model is not necessarily a good teacher in the knowledge distillation scenario}. Experimental results on CIFAR-100 and ImageNet show that EEKD method is significantly better than the state-of-the-art knowledge distillation methods. In particular, EEKD's performance even exceeds that of the standard ensemble distillation, which is another evidence that \emph{strong teachers do not necessarily produce strong students}.

The main contributions of our work are:
\begin{itemize}[itemsep= 0 pt,topsep = 0 pt, parsep = 0 pt]
	\item We propose a new experience ensemble knowledge distillation (EEKD) method to transfer the teacher's experience knowledge to the student. The proposed teacher's experience may be a supplement to the types of transferred knowledge in knowledge distillation.
	
	\item We find a surprising conclusion that strong ensemble teachers do not necessarily produce strong students, which may cause the rethinking of the route of ensemble distillation.

	\item Experimental results verify that EEKD exceeds the state-of-the-art knowledge distillation methods and the standard ensemble distillation method.
\end{itemize}

\section{Related work}
\label{gen_inst}

\subsection{Knowledge Distillation}

A vanilla knowledge distillation (KD, proposed in\cite{hinton2015distilling}) proposed to transfer some knowledge of a strong capacity teacher model to a compact student model by minimizing the Kullback-Leibler divergence between the soft targets of the two models. Since then, there have been plenty of work exploring variants of knowledge distillation. Fitnets \cite{romero2014fitnets} proposed to transfer the knowledge using both final outputs and intermediate ones. AT\cite{zagoruyko2016paying} proposed an attention-based method to match the activation-based and gradient-based spatial attention maps. FSP\cite{yim2017gift} proposed to explore the relationships between different layers, and compute the Gram matrix of feature maps across layers. CCKD\cite{peng2019correlation} proposed to transfer the correlation between input instances. CCM\cite{wang2021towards} proposed to match the class activation map of teacher with the class-agnostic activation map of student. In a survey of knowledge distillation \cite{gou2020knowledge}, they discussed different forms of knowledge in three categories, including response-based \cite{hinton2015distilling,meng2019conditional}, feature-based \cite{romero2014fitnets,zagoruyko2016paying,wang2021towards,ahn2019variational}, and relation-based \cite{yim2017gift,peng2019correlation,tung2019similarity,tian2019contrastive}. However, the transferred knowledge mentioned in the existing methods always comes from pre-trained teacher networks, ignoring the knowledge generated in the training process of the teacher, which we call \emph{teacher's experience}. It can be considered as a supplement to existing knowledge types used in knowledge distillation. Our proposed EEKD method aims to transfer the teacher's experience to the student network through ensemble learning technology.



\subsection{Ensemble Learning}

Neural network ensembles have been widely studied and applied in machine learning \cite{dietterich2000ensemble}, because an ensemble approach significantly boosts the prediction accuracy over the test set comparing to each individual model. Unfortunately, ensemble approaches always bring about the cost of higher computational power and memory, both training and testing cost. To reduce the testing cost, some work study the ensemble distillation, i.e., distilling the knowledge of an ensemble of teacher models to a student model. Existing approaches generally average the logits or soft labels of the multiple teacher models \cite{hinton2015distilling,zhang2018deep}. \cite{zhu2018knowledge} found that simply averaging the outputs would reduce the diversity, and generated the importance score corresponding to each model through the gate module. However, these methods rely on multi-teacher or multi-branch training, which can not effectively reduce the training cost.
To reduce the training cost, Snapshot Ensemble\cite{huang2017snapshot} proposed to train a single neural network with a cyclic learning rate strategy, in order to make the network converge to several local minima along its optimization path. Then they saved the intermediate models and averaged their predictions at test time. However, they did not validate the effectiveness in the knowledge distillation scenario. In this paper, we combined ensemble learning and knowledge distillation to reduce both training and testing costs. Different from \cite{huang2017snapshot}, our proposed EEKD method introduces a self-attention module to learn adaptive weights to stimulate the role of each intermediate model. In particular, compared with the standard ensemble distillation (integrating several full teacher models as an ensemble to distill a student model), EEKD shows stronger distillation performance and lower training cost simultaneously.

\subsection{Self-attention}

Attention was introduced in natural language processing (NLP) for encoding each word with others which are most relevant regarding to the target task\cite{bahdanau2014neural}. \cite{vaswani2017attention} proposed a transformer architecture based solely on attention mechanisms, which led to a big leap forward in capabilities for NLP tasks. Later, self-attention was also successfully used in many different forms of images tasks\cite{hu2018squeeze,bello2019attention,zhao2020exploring}. Vision Transformer (ViT)\cite{dosovitskiy2020image} used a sequence of embedded image patches as input to a standard transformer. It is the first convolution-free transformer that demonstrates comparable performance to CNN models. Now there are a number of variants of Transformers\cite{chen2021crossvit,touvron2021training,yan2021tbn}, pushing the computer vision task forward. In EEKD method, we adopt a self-attention module to learn the weights of different intermediate models automatically to improve the distillation performance.

\section{Method}
\label{headings}

In this section, we first briefly review the teacher-student knowledge distillation. Then, we describe the general framework of the proposed experience ensemble knowledge distillation (EEKD) method, analyze the main factors that limit its performance, and discuss the principles and techniques to improve its performance.

\subsection{Teacher-student Optimization}

Set a deep neural network as $\mathbb{M}:y=f\{x;\theta\}$, where $x$ denotes the input image, $y$ denotes the output of the network, and $\theta$ denotes the learnable parameters. These parameters are often initialized as random noise, and then optimized using a training set with $N$ data samples, $D =\left \{ \left ( x_{1},y_{1} \right ) , ... , \left ( x_{N},y_{N} \right )\right \}$. Conventional optimization algorithm works by sampling mini-batches or subsets from the training set. Each of them, denoted as $B$, is fed into the model to estimate the difference between prediction and ground-truth labels. The cross entropy loss function is as follows:
\begin{equation}
	L(B,\theta )=-\frac{1}{\left | B \right |}\sum_{(x_{n},y_{n})\in B}y_{n}^{T}\cdot logf(x_{n};\theta).
\end{equation}

The classic teacher-student knowledge distillation holds that the student network can obtain some valuable knowledge of the teacher network by imitating the output of the pre-trained teacher network. Teacher-student optimization was proposed\cite{hinton2015distilling}, in which an extra loss term was added to measure the KL divergence between teacher and student. The total loss function of the student is as follows:
\begin{equation}
	\begin{aligned}
		L^{s}(B,\theta^{s} )=-\frac{1}{\left | B \right |}\sum_{(x_{n},y_{n})\in B}\left \{ {\alpha y_{n}^{T}\cdot logf(x_{n};\theta^{s})} \right.  \\ \left. {+ (1-\alpha) KL\left [  f^{\tau}(x_{n};\theta^{t})||f^{\tau}(x_{n};\theta^{s})\right ] }\right \},
	\end{aligned}
	\label{ts}
\end{equation}
where $\theta^{t}$ and $\theta^{s}$ denote the parameters in teacher and student models respectively, $\tau$ is the temperature factor, $\alpha$ is a ratio parameter. The fitting goal of the student is to learn towards the teacher's output (a softened prediction) instead of the strict one-hot vector.

Despite the teacher-student optimization improves the performance of the student effectively, it only utilizes the knowledge of the pre-trained teacher and ignores the knowledge in the training process of the teacher. This means that the knowledge of the teacher has not been fully utilized, which is not efficient. It prompts us to propose the EEKD method, which explores the value of the experiential knowledge coming from the training process of the teacher. 

\subsection{Experience Ensemble Knowledge Distillation}

The key idea of EEKD is simple. 
The training process of the teacher model can be divided into $M$ stages uniformly. We can obtain $M$ intermediate models with a parameter set $\left \{  \theta_{1}^{t},\theta_{2}^{t},...,\theta_{M}^{t} \right \}$. The $M^{th}$ intermediate model is the full teacher model. Each intermediate model represents the capacity of the teacher model at a certain training stage, containing the teacher's perception of the data at the moment. We simply integrate the knowledge of all intermediate models by ensemble learning technique. That is, integrating the outputs of the intermediate models to get a stronger and more robust virtual teacher and then perform teacher-student optimization. Eq. \ref{ts} is transformed into the following form:

\begin{equation}
\begin{aligned}
L^{s}(B,\theta^{s} )=-\frac{1}{\left | B \right |}\sum_{(x_{n},y_{n})\in B}\left \{ {\alpha  y_{n}^{T} \cdot logf(x_{n};\theta^{s})} \right.  \\ \left. {+ (1-\alpha)  KL\left [(\sum_{i=1}^{M} w_{i}    f^{\tau}(x_{n};\theta_{i}^{t}))||f^{\tau}(x_{n};\theta^{s})\right ] }\right \},
\end{aligned}
\label{EEKD2}
\end{equation}
where $w_{i}$ denotes the weight coefficient of the $i^{th}$ intermediate model and satisfies:

\begin{equation}
\left\{
\begin{aligned}
0<w_{i}<1, \\ \sum_{i=1}^{M}w_{i}=1.
\end{aligned}
\right.
\label{EEKD1}
\end{equation}

Although the core idea of EEKD is simple, how to get a suitable virtual ensemble teacher is not trivial. Generally, excellent teachers are more likely to produce excellent students, so high-quality teacher models are preferred in knowledge distillation methods \cite{hinton2015distilling,romero2014fitnets,zagoruyko2016paying,yim2017gift,li2020online,chen2020online}. However, some recent studies argued a different view. \cite{mirzadeh2020improved} and \cite{gao2020residual} thought that a large model capacity gap between teacher and student may have a negative effect on knowledge transfer, and introduced assistant networks to narrow the gap. \cite{park2021learning} proposed to learn a student-friendly teacher by plugging in student branches during the training procedure. Therefore, different from ensemble learning, an optimal virtual ensemble teacher does not necessarily lead to the optimal performance of the student in knowledge distillation.

\subsection{Principles of Experience Ensemble Knowledge Distillation}
\label{sec33}

Comparing Equ.\ref{ts} and Equ.\ref{EEKD2}, the difference between EEKD and traditional methods is that the virtual ensemble teacher replaces the single teacher. $f(x_{n};\theta_{i}^{t})$, $w_{i}$ and $M$ are the main factors that directly affect the ability of the virtual ensemble teacher and the performance of EEKD. Respectively, $f(x_{n};\theta_{i}^{t})$ represents the performance of intermediate model, $w_{i}$ represents the weight of intermediate model, and $M$ represents the number of intermediate models. Therefore, we study the principles of EEKD from these three aspects.


\emph{1) Principle 1: The Quality of Intermediate Models}.

The quality of the intermediate models directly affects the performance of the ensemble teacher. In fact, training the same network model with different learning rate strategies will generate intermediate model sets with different quality. In the previous work, Snapshot Ensemble\cite{huang2017snapshot} investigated the impact of different intermediate model sets on the performance of ensemble model by using two classical learning rate strategies: normal cosine learning rate and cyclic cosine learning rate \cite{loshchilov2016sgdr}. Based on \cite{huang2017snapshot}, we further explore the distillation performance of different ensemble models. ResNet-110\cite{he2016deep} and WRN-40-2\cite{zagoruyko2016wide} are adopted as the architectures of teachers, while ResNet-20\cite{he2016deep} and WRN-40-1\cite{zagoruyko2016wide} are adopted as the architectures of students. For a fair comparison, we fix $M=5$ and $w_i=0.2$, $i\in\{1,2,...5\}$, train each teacher with normal cosine learning rate or cyclic cosine learning rate respectively for $200$ epochs, and save the intermediate teacher models at the $40^{th}$, $80^{th}$, $120^{th}$, $160^{th}$, $200^{th}$ epochs. Following \cite{huang2017snapshot}, we evaluate three types of teachers: ``Single Teacher", ``NoCycle Ensemble Teacher" and ``Cycle Ensemble Teacher". ``Single Teacher" is a teacher trained with normal cosine learning rate. ``NoCycle Ensemble Teacher" is an ensemble model of five intermediate teacher models uniformly extracted from the training process of ``Single Teacher". ``Cycle Ensemble Teacher" is an ensemble model of five intermediate models extracted from the training process of the teacher model, which is trained with cyclic cosine learning rate. More detailed experimental settings are shown in Sec. \ref{sec40}.

\begin{table}[h]
	\caption{Effects of different ensemble teachers on the performance of EEKD on CIFAR-100. ``ET" denotes ``Ensemble Teacher". Top 1 accuracy (\%) is averagely evaluated in three independent experiments. The best results for each column are \textbf{bold}.}
	\centering		\renewcommand{\arraystretch}{1.5}
	\setlength{\tabcolsep}{1mm}{
\begin{tabular}{|ccc|cc|c|}
	\hline
	\multicolumn{3}{|c|}{Teacher}                                                                           & \multicolumn{2}{c|}{Student}                                             & EEKD           \\ \hline
	\multicolumn{1}{|c|}{Network}                     & \multicolumn{1}{c|}{Ensemble type} & Accuracy       & \multicolumn{1}{c|}{Network}                    & Accuracy               & Accuracy       \\ \hline
	\multicolumn{1}{|c|}{\multirow{3}{*}{ResNet-110}} & \multicolumn{1}{c|}{Single Teacher}   & 73.41          & \multicolumn{1}{c|}{\multirow{3}{*}{ResNet-20}} & \multirow{3}{*}{68.91}  & 70.67           \\ \cline{2-3} \cline{6-6} 
	\multicolumn{1}{|c|}{}                            & \multicolumn{1}{c|}{NoCycle ET}    & 73.87          & \multicolumn{1}{c|}{}                           &                        & \textbf{72.60}  \\ \cline{2-3} \cline{6-6} 
	\multicolumn{1}{|c|}{}                            & \multicolumn{1}{c|}{Cycle ET}      & \textbf{75.56} & \multicolumn{1}{c|}{}                           &                        & 71.82           \\ \hline
	\multicolumn{1}{|c|}{\multirow{3}{*}{WRN-40-2}}   & \multicolumn{1}{c|}{Single Teacher}   & 76.53          & \multicolumn{1}{c|}{\multirow{3}{*}{WRN-40-1}}  & \multirow{3}{*}{70.38} & 72.68          \\ \cline{2-3} \cline{6-6} 
	\multicolumn{1}{|c|}{}                            & \multicolumn{1}{c|}{NoCycle ET}    & 77.18          & \multicolumn{1}{c|}{}                           &                        & \textbf{74.52} \\ \cline{2-3} \cline{6-6} 
	\multicolumn{1}{|c|}{}                            & \multicolumn{1}{c|}{Cycle ET}      & \textbf{78.39} & \multicolumn{1}{c|}{}                           &                        & 74.45          \\ \hline
\end{tabular}
	}
\label{table:principle1}
\end{table}


Table \ref{table:principle1} shows: 1) An ensemble Teacher of multiple intermediate teacher models has significantly better distillation performance than a single teacher, which means EEKD can easily surpass classical KD\cite{hinton2015distilling}. 2) ``Cycle Ensemble Teacher" has significantly better performance than ``NoCycle Ensemble Teacher", but the distillation performance of the former can not surpass the latter. This is another evidence that \emph{strong teachers do not necessarily produce strong students.} Further analysis shows that the higher performance of ``Cycle Ensemble Teacher" comes from the higher performance of each intermediate model and the higher diversity among intermediate models, especially the higher diversity. This has been verified in \cite{huang2017snapshot}. However, excessive diversity may lead to cognitive conflicts between multiple teachers, which makes it difficult for students to learn.

\emph{2) Principle 2: The Weights of Intermediate Models}.

In general ensemble models\cite{dietterich2000ensemble} and Snapshot Ensemble\cite{huang2017snapshot}, the equal weights are often adopted. In EEKD, different intermediate models contain the teacher's perception of the data at a certain training stage. An intuitive guess is that intermediate models in the late stages are more important than intermediate models in the early stages because they have higher performance, but such judgment may be too arbitrary. Therefore, we introduce a self-attention mechanism to automatically learn the weights of different intermediate models. It is an adaptive strategy aimed at getting the weight distribution that is beneficial to students' learning.



\textbf{Attention-based weights.} We represent the capacity of each model with the output of the last convolution layer, that is, a feature vector $u_{i}$ for the $i^{th}$ intermediate teacher model, a feature vector $v$ for the student model. Inspired by the self-attention mechanism\cite{vaswani2017attention}, we project the extracted feature vectors $u_{i}$ and $v$ into two subspaces separately by linear transformation:
\begin{equation}
E_{s}(v)=W_{s}^{T}\cdot v \ ; \  E_{t}(u_{i})=W_{t}^{T}\cdot u_{i}.
\label{attention}
\end{equation}
where $W_{s}^{T}$ and $W_{t}^{T}$ are learnable projection matrices of the student and teacher models. Similar to self-attention, $w_{i}$ is calculated as Embedded Gaussian distance with normalization:

\begin{equation}
w_{i}=\frac{e^{E_{s}(v)^{T}\cdot E_{t}(u_{i})}}{\sum_{j=1}^{M}e^{E_{s}(v)^{T}\cdot E_{t}(u_{j})}}.
\label{attention2}
\end{equation}

Compared with fixed weights, attention-based weights own the properties of \textbf{dynamic} and \textbf{diversity}. In the early training stage of the student, for the same input data, the drastic update of the parameters of the student network will lead to the dynamic change of the weights, so that the students can obtain more knowledge from the appropriate teachers adaptively. In the late training stage, the update of student network parameters is slow, but the weights of multiple intermediate teacher models are still diversified for different input data. 

To verify the effectiveness of attention-based weights, a comparative experiment is conducted. ResNet-110\cite{he2016deep} and WRN-40-2\cite{zagoruyko2016wide} are adopted as the architectures of teachers, while ResNet-20\cite{he2016deep} and WRN-40-1\cite{zagoruyko2016wide} are adopted as the architectures of students. For a fair comparison, we fix $M=5$, train each teacher with normal cosine learning rate for $200$ epochs, and save the intermediate teacher models at the $40^{th}$, $80^{th}$, $120^{th}$, $160^{th}$, $200^{th}$ epochs. The proposed attention-based weight strategy is evaluated by comparing with three fixed weight strategies, including mean, linear increase and linear decrease. More detailed experimental settings are shown in Sec. \ref{sec40}.

\begin{table}[h]
	\caption{Effects of different weight strategies on the performance of EEKD on CIFAR-100. Top 1 accuracy (\%) is averagely evaluated in three independent experiments. The best result for each column is \textbf{bold}.}
	\centering
	\renewcommand{\arraystretch}{1.5}
	\setlength{\tabcolsep}{1mm}{
		\begin{tabular}{|ccc|cc|c|}
			\hline
			\multicolumn{3}{|c|}{Teacher}                                                                           & \multicolumn{2}{c|}{Student}                                             & EEKD           \\ \hline
			\multicolumn{1}{|c|}{Network}                     & \multicolumn{1}{c|}{Weight strategy} & Accuracy       & \multicolumn{1}{c|}{Network}                    & Accuracy               & Accuracy       \\ \hline
			\multicolumn{1}{|c|}{\multirow{4}{*}{ResNet-110}} & \multicolumn{1}{c|}{Mean}   & 73.87          & \multicolumn{1}{c|}{\multirow{4}{*}{ResNet-20}} & \multirow{4}{*}{68.91}  & 72.60           \\ \cline{2-3} \cline{6-6} 
			\multicolumn{1}{|c|}{}                            & \multicolumn{1}{c|}{Linear increase}    & \textbf{74.21}          & \multicolumn{1}{c|}{}                           &                        & 70.79  \\ \cline{2-3} \cline{6-6} 
			\multicolumn{1}{|c|}{}                            & \multicolumn{1}{c|}{Linear decrease}      & 66.72 & \multicolumn{1}{c|}{}                           &                        & 69.51           \\ \cline{2-3} \cline{6-6} 
			\multicolumn{1}{|c|}{}                            & \multicolumn{1}{c|}{Attention-based}      & 74.15 & \multicolumn{1}{c|}{}                           &                        & \textbf{72.91}           \\ \hline
			\multicolumn{1}{|c|}{\multirow{4}{*}{WRN-40-2}}   & \multicolumn{1}{c|}{Mean}   & 77.18          & \multicolumn{1}{c|}{\multirow{4}{*}{WRN-40-1}}  & \multirow{4}{*}{70.38} & 74.52          \\ \cline{2-3} \cline{6-6} 
			\multicolumn{1}{|c|}{}                            & \multicolumn{1}{c|}{Linear increase}    & 77.32          & \multicolumn{1}{c|}{}                           &                        & 73.44 \\ \cline{2-3} \cline{6-6} 
			\multicolumn{1}{|c|}{}                            & \multicolumn{1}{c|}{Linear decrease}    & 70.81          & \multicolumn{1}{c|}{}                           &                        & 70.56 \\ \cline{2-3} \cline{6-6} 
			\multicolumn{1}{|c|}{}                            & \multicolumn{1}{c|}{Attention-based}      & \textbf{77.61} & \multicolumn{1}{c|}{}                           &                        & \textbf{74.78}          \\ \hline
		\end{tabular}
	}
	\label{table:principle2}
\end{table}

Table \ref{table:principle2} shows: 1) Compared with the other three weight strategies, attention-based weight strategy does not necessarily lead to the best ensemble performance, but can obtain the best distillation performance. 2) Compared with linear increase and linear decrease, the mean weight strategy is simple but effective. 3) Linear increase strategy does not achieve better distillation performance than mean strategy, indicating that intermediate models with higher performance are not necessarily worthy of assigning higher weights.

\emph{3) Principle 3: The Number of Intermediate Models}.

It is easy to save more intermediate models during the training process of teacher, but more intermediate models will significantly increase the ensemble teacher's inference cost. We explore the influence of the number of intermediate models on the performance and cost of knowledge distillation on CIFAR-100. WRN-40-2\cite{zagoruyko2016wide} is adopted as the architecture of teacher, while WRN-40-1\cite{zagoruyko2016wide} is adopted as the architecture of student. Normal cosine learning rate and the attention-based weight strategy are applicated. We evaluate the distillation performance and training cost in $M=1,3,5,7,10$ settings. Table \ref{table:principle3} shows that a large $M$ can improve the performance of EEKD. However, as $M$ gets larger, the training cost of the student significantly increases as a consequence of multiple intermediate teacher models doing inference. Linear cost increase is unacceptable especially for large models and large datasets. In practice, we need to make a trade-off between performance and cost.



\begin{table}[h]
	\caption{Effects of the number of intermediate models on the performance of EEKD on CIFAR-100. Top 1 accuracy (\%) is averagely evaluated in three independent experiments. Training cost of students is evaluated on TITAN Xp.}
	\begin{center}
		\renewcommand{\arraystretch}{1.5}
		\begin{tabular}{|c|c|c|c|c|c|}
			\hline
			Ensemble size    & 1     &   3   & 5     & 7    & 10   \\ \hline
			Test accuracy (\%)    & 72.68     &   74.02  & 74.78     & 74.89    & 74.98    \\ \hline
			Training cost (hour)    & 1.0    &   1.5   & 2.0     & 3.0    & 4.0   \\ \hline
		\end{tabular}
		
	\end{center}
	\label{table:principle3}
\end{table}

Based on the analysis of the above principles, we summarize the process of designing an excellent EEKD: 1) Train the teacher model with normal cosine learning rate, and extract a moderate number of intermediate models from the training process. 2) Adopt an attention-based weight strategy to make ensemble teacher model, then do the teacher-student distillation. Note that \emph{a high-performance ensemble teacher does not determine a high performance EEKD}.

\section{Experiments}

\label{others}
In this section, after the statement of datasets and experimental settings (Sec. \ref{sec40}), we evaluate the performance of EEKD by comparing it with state-of-the-art knowledge distillation methods in Sec. \ref{sec41}, and then compare EEKD with the standard ensemble distillation in Sec. \ref{sec42}.

\subsection{Datasets and Experimental Settings}
\label{sec40}
\begin{table*}[t]
	\caption{Comparison with state-of-the-art methods on CIFAR-100. Top 1 accuracy (\%) is averagely evaluated in three independent experiments. Partial results of the baseline methods are derived from \cite{tian2019contrastive}. ``-" denotes an unpublished result. The best result for each column is \textbf{bold}.}
	\label{table:exp1}
	\begin{center}
	\renewcommand{\arraystretch}{1.2}
\begin{tabular}{lcccccc}
	\hline
	Teacher & ResNet-110 & ResNet-110 & ResNet-110  & WRN-40-2 & WRN-40-2 & WRN-40-2    \\
	Student & ResNet-20  & ResNet-32  & MobileNetV2 & WRN-40-1 & WRN-16-2 & MobileNetV2 \\ \hline
	Teacher & 73.41      & 73.41      & 73.41       & 76.53    & 76.53    & 76.53       \\
	Student & 68.91      & 70.16      & 64.49       & 71.95    & 73.56    & 64.49       \\ \hline
	KD\cite{hinton2015distilling}      & 70.67      & 72.48      & 68.63       & 72.68    & 74.92    & 68.03       \\
	FitNet\cite{romero2014fitnets}  & 70.67      & 71.06           &   -          & 72.94    & 75.12    & -            \\
	AT\cite{zagoruyko2016paying}      & 70.91      & 72.31           & -            & 72.94    & 75.32    &  -           \\
	SP\cite{tung2019similarity}      & 71.02      & 72.69           & -            & 73.18    & 74.98    &  -           \\
	CCKD\cite{peng2019correlation}    & 70.88      & 71.48           & -            & 72.22    & 75.00         & -            \\
	VID\cite{ahn2019variational}     & 71.10      & 72.61           & -            & 72.52    & 75.14    &    -         \\
	CRD\cite{tian2019contrastive}     & 71.46      & 73.48           & -            & 73.59    & 75.48    & -            \\
	TAKD\cite{mirzadeh2020improved}    & 71.42      & -           & -            & 73.52    & 76.04         &  -           \\ \hline
	EEKD$_{mean,M=5}$    & 72.60      &  73.65          &  70.78           & 74.52    & 76.57      &  71.04            \\
	EEKD$_{at,M=5}$    & 72.91      & 74.02          &  70.83           & 74.78    & \textbf{76.82}       & 71.89            \\
	EEKD$_{at,M=10}$    & \textbf{73.23}      & \textbf{74.34}           &  \textbf{71.11}         & \textbf{74.98}    & 76.76      & \textbf{72.21}       \\ \hline
\end{tabular}
	\end{center}
\end{table*}


We evaluate our EEKD method on two well calibrated image classification datasets CIFAR-100\cite{krizhevsky2009learning} and ImageNet-1K\cite{deng2009imagenet}. The CIFAR-100 dataset contains 50,000 training images with 500 images per class and 10,000 test images with 100 images per class. It comprises 32 $\times$ 32 pixel RGB images with 100 classes. We follow the standard augmentation in\cite{howard2013some}. That is, the training images are padded 4 pixels, and then randomly clipped to $32\times32$ combined with random horizontal flipping. The original 32 $\times$ 32 pixel images are used for testing. ImageNet-1K\cite{deng2009imagenet} dataset contains 1.2 million training images and 50,000 validation images of 1000 categories. We adopt the same augmentation strategy as\cite{yun2019cutmix} and apply a center cropping in testing. 

To investigate the generalization ability of our method, we construct teacher-student pairs with similar structures (e.g., ResNet-110/ResNet-20, WRN-40-2/WRN-40-1) and different structures (e.g., ResNet-110/MobileNetV2). On CIFAR-100, we run a total of $200$ epochs for all methods with SGD optimizer. We set batchsize to $64$, momentum to $0.9$, and weight decay to $5e^{-4}$. We adopt cosine learning rate strategy and initialize the learning rate to $0.1$. On ImageNet-1K, we run a total of 90 epochs, set batchsize to 256. We set the hyperparameter temperature as 4 for all methods, other parameters following the original papers. For EEKD, unless otherwise specified, we always set $\alpha = 0.5$, $M=5$, $\tau=5$, and adopt the attention-based weight strategy.

\subsection{Comparison with the State-of-the-art}
\label{sec41}

\textbf{Results on CIFAR-100.} For extensive experiments, we adopt ResNet-110\cite{he2016deep} and WRN-40-2\cite{zagoruyko2016wide} as the architectures of teachers, while ResNet-20\cite{he2016deep}, ResNet-32\cite{he2016deep}, WRN-40-1\cite{zagoruyko2016wide}, WRN-16-2\cite{zagoruyko2016wide}, and MobileNetV2 (width multiplier 0.5)\cite{howard2017mobilenets} as the architectures of students. We investigate a large number of baseline methods in recent years. For EEKD, we show the results on three different setting, mean weights and $M=5$, attention-based weights and $M=5$, attention-based weights and $M=10$. As shown in table \ref{table:exp1}, we can have the following observations. 1) EEKD easily exceeds all baseline methods by a wide margin, which shows that EEKD has an absolute advantage over baseline methods. 2) EEKD significantly narrows the gap between teachers and students, improves students' accuracy by 4.86\% on average. In some settings, the student's performance even exceeds the teacher's performance (e.g., ResNet-110/ResNet-32 and WRN-40-2/WRN-16-2). Consistent conclusions indicate that directly adding teachers' experiential knowledge is more effective than skillfully changing the type of knowledge representation based on a pre-trained teacher.

\textbf{Results on ImageNet.}
For a fair comparison, following \cite{peng2019correlation,liu2021improved}, we adopt the ResNet-50\cite{he2016deep} and ResNet-34\cite{he2016deep} as the architectures of teachers, MobileNetV2 with 0.5 width multiplier and ResNet-18\cite{he2016deep} as the architectures of students. Table \ref{table:exp2} shows that the proposed EEKD achieves the highest accuracy. Specifically, For ResNet-34/ResNet-18, it reduces the performance gap between the teacher and the student from 3.56\% to 1.64\% , a 54\% relative improvement. For ResNet-50/MobileNetV2, it reduces the performance gap between the teacher and the student from 11.3\% to 6.7\% , a 41\% relative improvement. These results demonstrate the superiority of our proposed EEKD. 
\begin{table}[h]
	\caption{Comparison with state-of-the-art methods on ImageNet-1K. Top 1 accuracy (\%) is averagely evaluated in three independent experiments. Results of the baseline methods are derived from \cite{peng2019correlation} and \cite{liu2021improved}. ``-" denotes an unpublished result. The results of EEKD are \textbf{bold}.}
	\label{table:exp2}
	\begin{center}
		\renewcommand{\arraystretch}{1.2}
	\begin{tabular}{lcc}
		\hline
		Teacher & ResNet-50   & ResNet-34 \\
		Student & MobileNetV2 & ResNet-18 \\ \hline
		Teacher & 75.5        & 73.31     \\
		Student & 64.2        & 69.75     \\ \hline
		KD\cite{hinton2015distilling}      & 66.7        & 70.66     \\
		AT\cite{zagoruyko2016paying}      & 65.4        & 70.70     \\
		SP\cite{tung2019similarity}      & -           & 70.62     \\
		CCKD\cite{peng2019correlation}    & 67.7        & 69.96     \\
		CRD\cite{tian2019contrastive}     & -           & 71.17     \\
		ACKD\cite{liu2021improved}    & -           & 71.33     \\ \hline
		EEKD    & \textbf{68.8}       & \textbf{71.67}\\ \hline   
	\end{tabular}
\end{center}
\end{table}

\subsection{Comparison with Standard Ensemble Distillation}
\label{sec42}

EEKD is to obtain $M$ intermediate models in the training process of one teacher model. The standard ensemble distillation (SED) is to train $M$ full teacher models independently and average their output to distill a single student model. \cite{huang2017snapshot} verified that a standard ensemble certainly has more powerful performance due to higher individual accuracy and greater diversity. So, it is natural to guess that the standard ensemble distillation should have better distillation performance than EEKD. We evaluate the distillation performance and training cost of two methods in $M=1,3,5,7,10$ settings on CIFAR-100. We adopt WRN-40-2\cite{zagoruyko2016wide} as the architecture of teacher, WRN-40-1\cite{zagoruyko2016wide} as the architecture of student. Table \ref{table:exp3} shows a surprising conclusion. Compared with SED, EEKD not only has less training cost, but also has significantly better distillation performance. For example, when $M=7$, EEKD outperforms SED by 1.21\% but consumes only 40\% cost of SED. It once again confirms that high-performance ensemble teacher does not necessarily lead to high-performance students, which is the same as the conclusion of Sec. \ref{sec33}.

\begin{table}[h]
	\caption{Comparison with standard ensemble distillation (SED) on CIFAR-100. Top 1 accuracy (\%) is averagely evaluated in three independent experiments. Total training cost (hour) consists of training teachers and distilling students.}
	\label{table:exp3}
	\begin{center}
		\renewcommand{\arraystretch}{1.2}
\begin{tabular}{llccccc}
	\hline
	\multicolumn{2}{l}{Ensemble size}     & 1     & 3     & 5     & 7     & 10    \\ \hline
	\multirow{2}{*}{Accuracy}      & SED  & 72.68 & 73.21 & 73.46 & 73.68 & 73.55 \\
	& EEKD & 72.68 & 74.02 & 74.78 & 74.89 & 74.98 \\ \hline
	\multirow{2}{*}{Training cost} & SED  & 2.0   & 3.7   & 4.3   & 10.3  & 13.8  \\
	& EEKD & 2.0   & 2.4   & 3.2   & 4.1   & 5.2   \\ \hline
\end{tabular}
\end{center}
\end{table}



\section{Conclusion}

In this paper, we have proposed a simple but efficient knowledge distillation method, experience ensemble knowledge distillation (EEKD). We first extract abundant useful knowledge from intermediate models of the teacher network, and then adaptively integrate the experience knowledge with an attention-based weight module, finally transfer the integrated knowledge to the student model. The experimental results show that EEKD achieves the new state-of-the-art in CIFAR-100 and ImageNet. In particular, EEKD even surpasses the standard ensemble distillation on the premise of saving training cost, which may cause a rethinking of the way of knowledge distillation using ensemble teachers. Next, we will continue to study the deep theory of EEKD and explore the essential role of teachers' experience in knowledge distillation. Furthermore, the constructed strong and robust ensemble teacher in EEKD can be used in combination with other methods to further improve their performance, which will further verify the usefulness of teachers' experience.  

{\small
	\bibliographystyle{ieee_fullname}
	\bibliography{egbib}
}

\end{document}